\pdfoutput=1
\documentclass[11pt]{article}

\usepackage{acl}

\usepackage{times}
\usepackage{latexsym}
\usepackage{natbib}
\usepackage{graphicx}
\usepackage{multirow}
\usepackage{amsmath}
\usepackage{amsfonts}
\usepackage{amssymb,amsfonts}
\usepackage{xcolor}
\usepackage{bbm}
\usepackage{booktabs}
\usepackage{algorithm, algorithmic}
\usepackage{subcaption}
\usepackage{soul}
\usepackage{color, xcolor}
\usepackage{hyperref}
\usepackage{pifont}
\usepackage{lipsum}

\definecolor{c1}{HTML}{AEC670}
\definecolor{c2}{HTML}{EDC5AB}

\usepackage[T1]{fontenc}

\usepackage[utf8]{inputenc}

\usepackage{microtype}

\usepackage{inconsolata}

\title{PokeMQA: Programmable knowledge editing for Multi-hop Question Answering}

\author{ Hengrui Gu\textsuperscript{1}, Kaixiong Zhou\textsuperscript{2}, Xiaotian Han\textsuperscript{3}, Ninghao Liu\textsuperscript{4}, \\ {\bf Ruobing Wang\textsuperscript{1}, Xin Wang\textsuperscript{1\ding{66}}}\\
$^{1}$School of Artificial Intelligence, Jilin University \\
$^{2}$ Department of Electrical and Computer Engineering, North Carolina State University \\
$^{3}$ Department of Computer Science and Engineering, Texas A\&M University\\
$^{4}$School of Computing, University of Georgia \\}

\begin{document}
\maketitle
\newcommand\blfootnote[1]{%
\begingroup
\renewcommand\thefootnote{}\footnote{#1}%
\addtocounter{footnote}{-1}%
\endgroup
}
\begin{abstract}
Multi-hop question answering (MQA) is one of the challenging tasks to evaluate machine's comprehension and reasoning abilities, where large language models (LLMs) have widely achieved the human-comparable performance. Due to the dynamics of knowledge facts in real world, knowledge editing has been explored to update model with the up-to-date facts while avoiding expensive re-training or fine-tuning. Starting from the edited fact, the updated model needs to provide cascading changes in the chain of MQA. The previous art simply adopts a mix-up prompt to instruct LLMs conducting multiple reasoning tasks sequentially, including question decomposition, answer generation, and conflict checking via comparing with edited facts. However, the coupling of these functionally-diverse reasoning tasks inhibits LLMs' advantages in comprehending and answering questions while disturbing them with the unskilled task of conflict checking. We thus propose a framework, \underline{P}r\underline{o}grammable \underline{k}nowledge \underline{e}diting for \underline{M}ulti-hop \underline{Q}uestion \underline{A}nswering (PokeMQA), to decouple the jobs. Specifically, we prompt LLMs to decompose knowledge-augmented multi-hop question, while interacting with a detached trainable scope detector to modulate LLMs behavior depending on external conflict signal. The experiments on three LLM backbones and two benchmark datasets validate our superiority in knowledge editing of MQA, outperforming all competitors by a large margin in almost all settings and consistently producing reliable reasoning process. Our code is available at 
\href{https://github.com/Hengrui-Gu/PokeMQA}{https://github.com/Hengrui-Gu/PokeMQA}.\blfootnote{\textsuperscript{\ding{66}} Corresponding author}

\end{abstract}

\section{Introduction}



\label{intro}
\begin{figure}[t]
\centering
\includegraphics[width=0.48\textwidth]{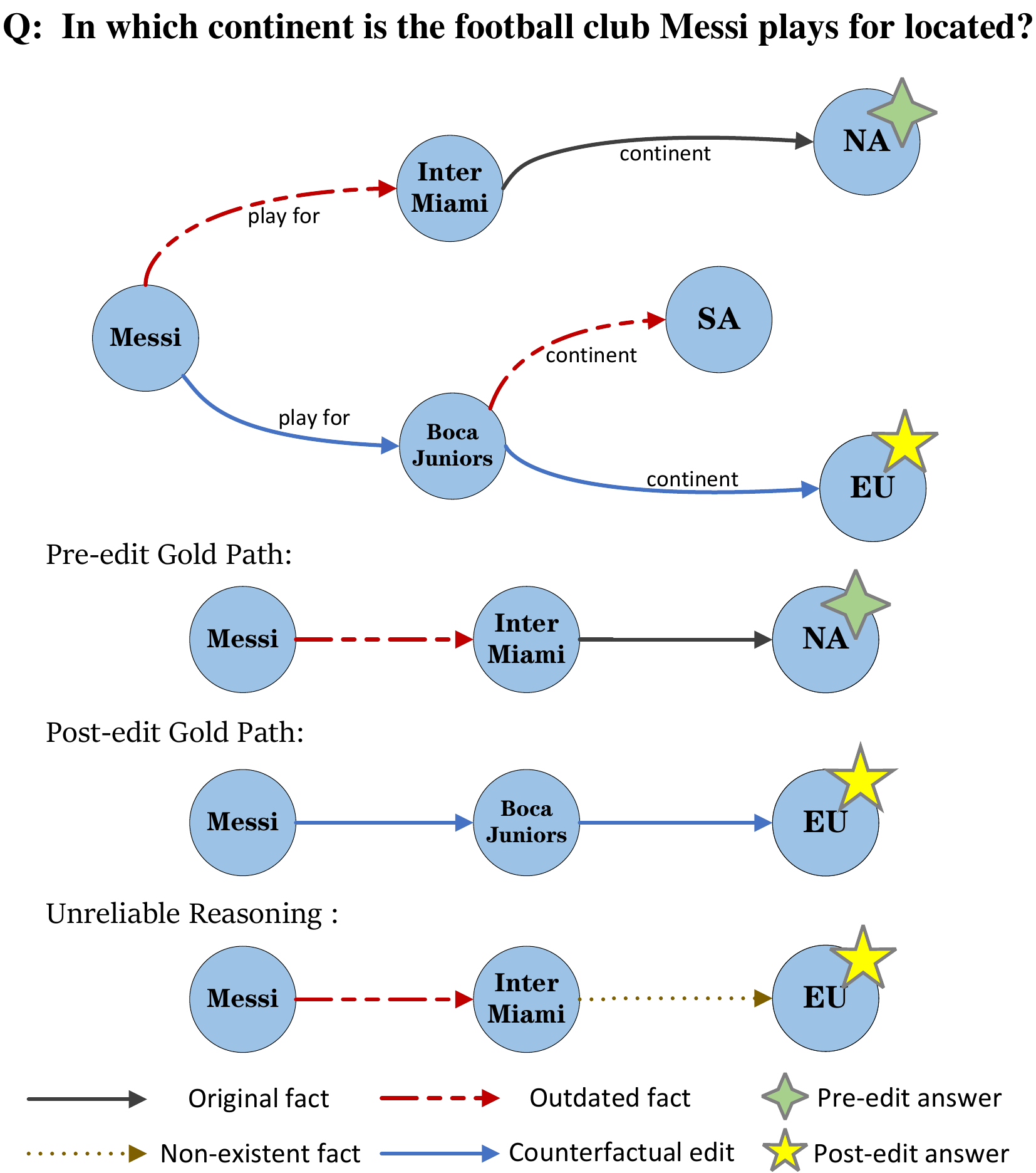}
\caption{An example of multi-hop question answering under knowledge editing, which consists of relevant knowledge facts and three specific reasoning paths solving the two-hop question. For the unreliable reasoning, it uses a outdated and a non-existent fact and end up with the right answer \textit{Europe}.}
\label{unreliable_reasoning}
\end{figure}

Multi-hop question answering (MQA) requires a sequence of interacted knowledge facts to reach the final answer. 
For instance, considering the two-hop question in Figure \ref{unreliable_reasoning}, it is necessary to deduce the intermediate answer \textit{Inter Miami} through the fact "Messi plays for Inter Miami", and then deduce the final answer \textit{NA} through another fact "Inter Miami is located in North America"). MQA poses a great challenge to reasoning abilities of question answering systems (\citealp{multihop1}, \citealp{multihop2}, \citealp{multihop3}). Thanks to the natural language comprehending and reasoning brought by large-scale pre-training, large language models (LLMs) have proven its indispensable utility in MQA tasks (\citealp{LLMqa}, \citealp{LLMqa2}, \citealp{compressllm1}, \citealp{kgllm3}).

However, the knowledge within LLMs may be factually wrong or become invalid over time. 
To ensure the correctness of LLMs without necessitating expensive retraining (\citealp{memoryllm2}), technique of knowledge editing has been carried out to provide efficient and targeted updates on model behaviors (\citealp{enn}, \citealp{ft-zhu}, \citealp{knowledgeeditor} ). There are two popular approaches: parameter-modification based editing and memory-based editing. The former one modifies the internal model weights according to edited facts  through meta-learning, fine-tuning, or knowledge locating (\citealp{rome}, \citealp{mend}, \citealp{memit}). The latter approach leverages an external memory to explicitly store the edited facts (or termed as edits) and reason over them, while leaving LLMs parameters unchanged \citep{serac, ike}. Memory-based model editing is generally adopted due to its simpleness and agnostic to backbone LLMs. 



In the context of MQA, MeLLo \citep{mello} is first proposed 
by designing a multipurpose prompt to instruct LLMs conducting the reasoning tasks of question decomposition and knowledge editing sequentially. In particular, after decomposing the multi-hop questions, LLMs generate a tentative answer for each subquestion and then detect whether there exists factual conflict between tentative answer and edited facts in memory (e.g., statements of "the current British Prime Minister is Rishi Sunak" and "the current Prime Minister of the UK is Liz Truss" are factually incompatible with each other). By repeatedly prompting LLMs, MeLLo reaches the answer of multi-hop question.




However, the coupling of question decomposition and knowledge editing imposes considerable demands on LLMs to precisely perform reasoning as demonstrations in context. \underline{\textit{First}}, the knowledge editing requires LLMs to fully understand the semantics of two candidate facts and then make conflict detection based on the factual compatibility between them. In the few-shot prompting, LLMs are prone to underfit this editing logic due to inadequate supervision signals, especially when embedded in a more complex task \citep{decomposedprompting}, i.e. question decomposition. \underline{\textit{Second}}, within a unified prompt, the incorporation of knowledge editing instruction introduces noise to question decomposition in the similar way.
Such superposed noise prevents LLMs from fully focusing on parsing the syntactic structure of multi-hop questions to precisely identify the subquestions.


Thus, we propose \underline{P}r\underline{o}grammable \underline{k}nowledge \underline{e}diting for \underline{M}ulti-hop \underline{Q}uestion \underline{A}nswering (PokeMQA), where we decouple the two essential tasks, i.e. question decomposition and knowledge editing, to alleviate burdens on LLMs while introducing auxiliary knowledge prompt to assist question decomposition. 
Specifically, we offload the conflict detection in knowledge editing with a programmable scope detector, which is used to detect whether a subquestion lies within the scope affected by any edited facts in semantic space (\underline{\textit{Challenge \#1}}).
A two-stage scope detector is designed: In pre-detection stage, we efficiently filter out a substantial number of irrelevant edits; In conflict-disambiguation stage, we perform precise retrieval on the remaining few candidate edits. Our two-stage framework provides both computational efficiency and expressiveness given the high volume of edited facts in real scenarios. The retrieved edits are used to calibrate LLMs behavior.
Moreover, we propose a knowledge prompt to augment parse analysis in the process of question decomposition (\underline{\textit{Challenge \#2}}). The knowledge prompt recognizes key entity from input question and retrieves its external information from a knowledge source to trig the correct decomposition. 





Additionally, we observe that the multi-hop question answering process may use the outdated or non-existent facts, but occasionally ends up with the right answer. We refer to this situation as unreliable reasoning (as shown in Figure \ref{unreliable_reasoning}). In order to faithfully evaluate models' reasoning ability,
we propose a new metric called hop-wise answering accuracy (Hop-Acc), measuring the extent how LLMs follow demonstrations, conduct question decomposition step by step, and generate desired answer to each step towards solving the multi-hop question. 

\section{Multi-hop Question Answering under Knowledge Editing}



\label{hop-acc}
\noindent \textbf{Notations.  }Following previous work (\citealp{mello,rome}), we denote a fact as a triplet $\left ( s,r,o\right )$, consisting of the subject $s$, object $o$, and relation $r$ between them, such as (\textit{Messi}, \textit{play for}, \textit{Inter Miami}). An edited fact (i.e., edit) is the knowledge fact that we want to update and is represented in the same form $\left ( s,r,o\right )$, such as (\textit{Messi}, \textit{play for}, \textit{Boca Juniors}). We consider a multi-hop question $Q$, where answering $Q$ requires sequentially querying and retrieving multiple facts. These facts are presented in the order they were queried, forming a \textit{chain of facts} $\left \langle \left ( s_{1},r_{1},o_{1}\right ),\dots ,\left ( s_{n},r_{n},o_{n}\right )\right \rangle$, where $s_{i+1}=o_{i}$ and $o_{n}$ is the final answer, which uniquely represents an inter-entity path $\mathcal{P} =  \left \langle s_{1},o_{1},\dots ,o_{n}\right \rangle$. It should be noted that except for $s_{1}$, all the other entities $o_{1},\dots ,o_{n}$ in $\mathcal{P}$ do not appear in $Q$ and need to be deduced either explicitly or implicitly through factual reasoning (like \textit{Inter Miami} and \textit{North America} in the multi-hop question in Figure~\ref{unreliable_reasoning}). 
If we replace the invalid fact $\left ( s_{i},r_{i},o_{i}\right )$ with edit $e=\left ( s_{i},r_{i},o_{i}^{*}\right )$ in a multi-hop question, 
due to the cascading effect caused by the edited fact, the chain of facts accordingly changes to $\left \langle \left ( s_{1},r_{1},o_{1}\right ),\dots ,\left ( s_{i},r_{i},o_{i}^{*}\right ),\dots\left ( s_{n}^{*},r_{n},o_{n}^{*}\right )\right \rangle$. The updated inter-entity path is  $\mathcal{P}^{*} =  \left \langle s_{1},o_{1},\dots,o_{i}^{*} ,\dots,o_{n}^{*}\right \rangle$, which indicates the reasoning path to the final answer of $Q$ has changed after being influenced by edit $e$.


\noindent \textbf{MQA under knowledge editing.} Given a set of edits $\mathcal{E}=\left \{ e_{1},\dots,e_{m} \right \}$ and a language model $f$ to be edited, for a multi-hop question $Q$, its inter-entity path becomes $\mathcal{P}^{*} =  \left \langle s_{1},o_{1}^{*},\dots ,o_{n}^{*}\right \rangle$ after being affected by edits in $\mathcal{E}$. 
The goal of multi-hop question answering under knowledge editing can be formally described as producing an edited language model $f_{\mathrm{edit}}$ conditioned on $f$ and $\mathcal{E}$, which can deduce the inter-entity path $\mathcal{P}^{*}$ and finally output the post-edit answer $o_{n}^{*}$ to question $Q$. We denote $\mathcal{P}^{*}$ as \textit{gold path} of $Q$ (as shown in Figure~\ref{unreliable_reasoning}).
Different from the previous work, we not only evaluate whether edited model $f_{\mathrm{edit}}$ output the desired final answers, but also check the correctness of their intermediate reasoning paths, providing faithful MQA performance results for knowledge editing.

\noindent \textbf{Edit scope.} In line with our work, we make some modifications to this concept that was originally proposed by \citep{serac}. For an edit $e=\left ( s,r,o\right )$, we define the single-hop question $q$ describing $(s,r)$ with the answer being $o$ as its \textit{atomic question}. It should be noted that the atomic question corresponding to a specific edit is not unique but rather a set of semantically equivalent questions (e.g., "What is the country of origin of hockey?" and "Where did hockey originate?"). We refer to the set as the \textit{scope} of an edit, denoted as $S(e)$. After making an edit $e=\left ( s,r,o\right )$, the answers to those questions in $S(e)$ should change to $o$ accordingly. Compared with the previous work, we define the edit scope based on the unit of atomic question, excluding the original multi-hop question, which typically has a much more complex syntactic structure. This simplified definition facilitates the programmable scope detector to learn the semantic patterns represented by $S(e)$ and then make precise edit retrieval to adjust LLMs behavior.



\begin{figure*}[t]
\centering
\centerline{\includegraphics[width=\textwidth]{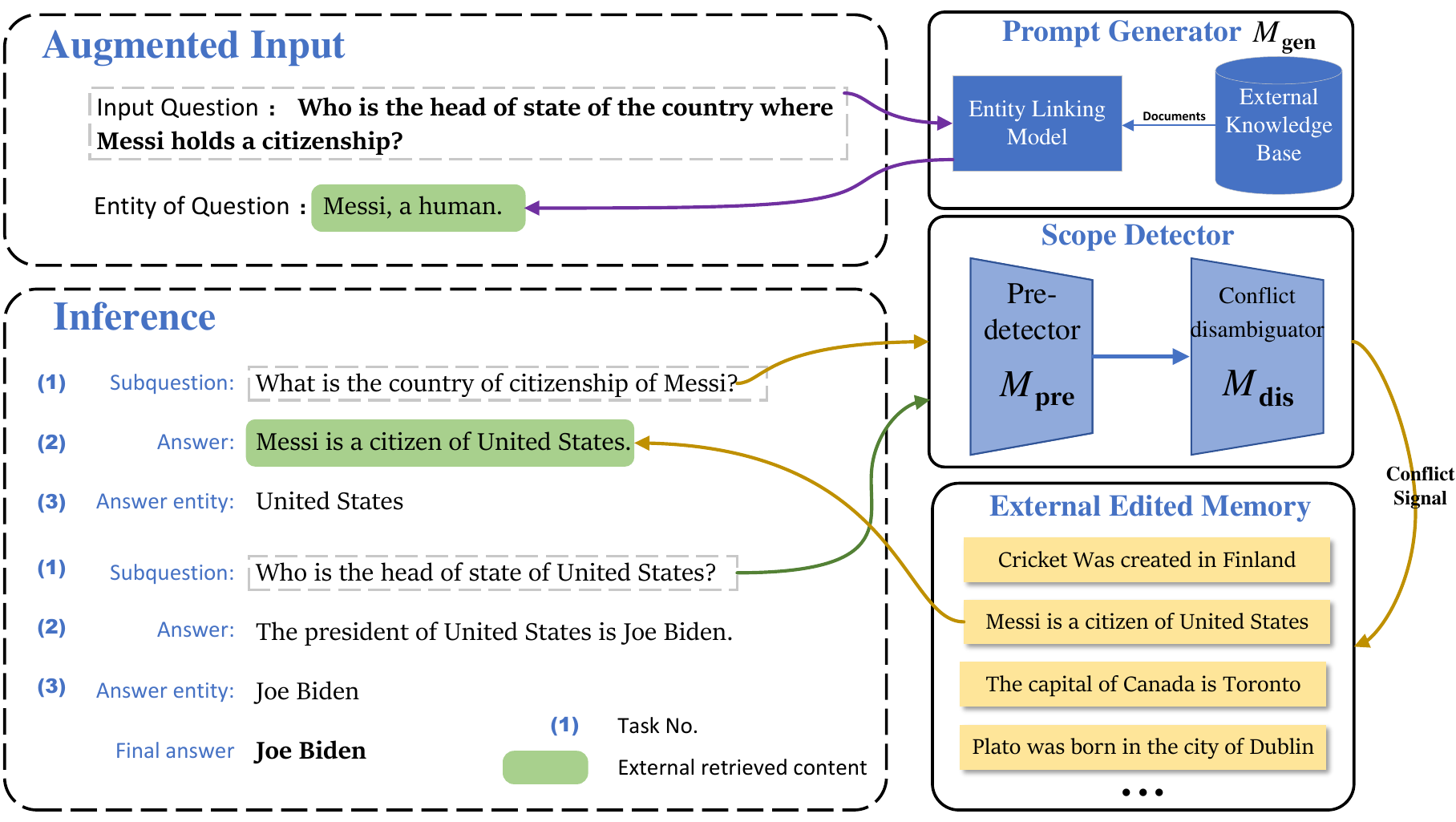}}
\caption{The illustration of our proposed method PokeMQA. PokeMQA  leverages external knowledge base to construct knowledge prompt, facilitating the decomposition of the first subquestion. It then alternately executes subsequent question decomposition, knowledge editing with programmable scope detectors, and answer generation for MQA.
The concrete prompts used in PokeMQA are shown in Appendix \ref{promptinPMQA}. }
\label{pokemqa_framework}
\vspace{-7pt}
\end{figure*}

\section{Programmable Editing in Memory of Multi-hop Question Answering}
\label{method}

\subsection{Workflow of PokeMQA}
\label{editinfer}
As illustrated in Figure~\ref{pokemqa_framework}, PokeMQA is a lightweight model editor that can be seamlessly integrated into any backbone LLMs, without changing parameters in the deployed language models. 
This empowers the language models to be robust to respond to questions based on edited facts. From initiating the editor to successfully addressing a question, the proposed procedure involves two steps as follows:

\noindent \textbf{Storing edits in memory.}
When receiving a set of edits $\mathcal{E}=\left \{ e_{1},\dots ,e_{m}\right \}$, PokeMQA first uses manually-defined template to convert each edit triplet $e$ into a natural language statement $t$ (as in \citealp{mello}), then explicitly stores them in an external memory $\mathcal{M}=\left \{ t_{1},\dots ,t_{m}\right \}$ for query and retrieval.

\noindent \textbf{Inference by checking with edit memory. }
Considering an input of multi-hop question, we adopt in-context learning \citep{icl} and provide a few demonstrations (i.e., input-label pairs) as the few-shot prompt to teach models to execute the following three tasks 
alternately: I) Identify the next subquestion (i.e., atomic question) conditioned on the input question and current inference state in LLMs; II) Detect whether this subquestion falls within the edit scope and generate answer; III) Extract the answer entity for this subquestion in LLMs. Note that this answer entity is either used to decompose the next subquestion at Step I or released as the final answer.

Particularly, we propose the programmable scope detector to detach the knowledge editing task in Step II from LLMs. Previous work~\citep{mello} generates tentative answers for each subquestion and checks the semantic conflict between tentative answer and retrieved edit in LLMs. With the slight supervision signals from the few-shot prompt, it is challenging for LLMs to compare their semantic patterns and make the correct conflict detection. In this work, the proposed scope detector takes the subquestion as input and detects whether it falls within the scope of any edit in $\mathcal{M}$. If so, the detector sends the factual conflict signal back with a chosen edit statement. The statement serves as a prompt to instruct LLMs to infer the answer from the edit. Otherwise, the factual conflict signal is empty and LLMs directly generate answer based on their internal knowledge. 


In addition, we propose knowledge prompt to correct the question decomposition in Step I. In MQA, identifying the leading subquestion (i.e., the first decomposed subquestion) might be challenging due to insufficient contextual information. Specifically, given the input of multi-hop question, one lacks the explicit question entity and entity-related fact, which are available when identifying the subsequent subquestions. To address this, we innovatively employ the knowledge prompt generator to preprocess the input question. It recognizes the key entity and retrieves relevant documents from an external knowledge base to create a knowledge prompt. Then, we concatenate the input question and knowledge prompt to form an augmented input, effectively resolving the issue.


Owing to the proposed scope detector and knowledge prompt, PokeMQA allows language models to focus on question decomposing and answering, formulating a reliable reasoning path. The details of the proposed components are stated below.


\subsection{Programmable Scope Detector}
\label{sub1}

Motivated by \citep{serac}, we utilize a programmable scope detector for conflict detection and design a task-specific training approach to identify effective edit scope patterns.

\noindent \textbf{Architectures. }The scope detector can be formally described as $g(t,q):\mathcal{T} \times  \mathcal{Q}\to \left [ 0,1\right ]$, which predicts the probability that an atomic question $q$ falls into the scope of the edit statement $t$ (in terms of the edit $e$). The scope detector can be implemented as arbitrary text classification models (\citealp{textclass}, \citealp{textclass2}, \citealp{classification4}, \citealp{classification5}). In our framework, considering both expressiveness and computational efficiency, we choose two lightweight, yet complementary models. The two models are denoted as $g_{\phi }$ and $g_{\psi }$, respectively. For an input pair $(t,q)$, $g_{\phi }$ calculates the embeddings for $t$ and $q$ separately and models the log-likelihood by the negative squared Euclidean distance in the embedding space. Model $g_{\psi }$ concatenates $t$ and $q$ together as a unified input for the sequence classification task.


In our framework, the models $g_{\phi}$ and $g_{\psi}$ serve as pre-detector $M_{\mathrm{pre}}$ and conflict disambiguator $M_{\mathrm{dis}}$, respectively. We combine them together to establish a \textit{two-stage edited fact retrieval} framework. The pre-detector filters out the enormous semantically irrelevant edits from memory efficiently, while the conflict disambiguator accurately locates on candidate edit with the highest likelihood.
The details are given in Appendix \ref{twostage}. Once the detector finally believes that an input atomic question falls into the scope of any edit in $\mathcal{M}$, it retrieves the edited statement of candidate edit 
and sends it back along with a factual conflict signal to guide the language model generation process.


\noindent \textbf{Training scope detector.    }
According to edit memory $\mathcal{M}=\left \{ t_{1},\dots ,t_{m}\right \}$, we 
build up a training dataset $\mathcal{D}_{\mathrm{train}}=\left \{ \left ( t_{1},q_{1}\right ),\dots ,\left ( t_{m},q_{m}\right )\right \}$. See Appendix \ref{datasetconstruction} for more details on the construction of the dataset. To learn the scope covered by each edit statement $t_i$, we use a binary cross-entropy loss with negative sampling \citep{negsam1} as the training objective:
\begin{equation}\label{bce}
        \mathcal{L}=-\log g(t_{i},q_{i})-\mathbb{E}_{q_{n} \sim P_{n}(q)}\left[\log(1-g(t_{i},q_{n}))\right ], 
\end{equation}
\noindent where $P_{n}$ is a negative sampling distribution and we set it to a uniform distribution over each minibatch. Note that $M_{\mathrm{pre}}$ and $M_{\mathrm{dis}}$ are trained separately using the above supervised learning setting.

\noindent \textbf{Model selection.   } 
In practice, we observed that using the traditional classification metric (accuracy) to validate the detector's performance can often result in underfitting. We believe this is due to the unique characteristics of the conflict detection task. Thus, we define two novel task-specific metrics to select detector models and guide early stopping during training: \textit{Success Rate} and \textit{Block Rate}. The \textit{Success Rate} measures the accuracy to retrieve the correct edit statement $t_i$ for a target question $q_i$ from a set of candidates:
\begin{equation}\label{success rate}
    SR=\frac{1}{N}\displaystyle\sum_{i=1}^{N} \mathbbm{1} \left [ \bigwedge_{(t,q)\in\mathcal{D}_{val}}(g(t_{i},q_{i})\ge  g(t,q_{i}) )\right ],
\end{equation}
\noindent where $\mathbbm{1}\left ( \cdot \right )$ is the indicator function, $N$ is the size of validation set $\mathcal{D}_{val}$, and $\bigwedge$ denotes the $\texttt{AND}$ gate. For the target pair $(t_{i},q_{i})$, the retrieval is precise if and only if its detection likelihood is higher than the other pairs $(t,q_{i})$, which are synthesized by replacing the target edit statement $t_i$ with candidates from  $\mathcal{D}_{val}$. On the other hand, metric \textit{Block Rate}  quantifies the extent of detector models to inhibit the unrelated edit statements for a target question:

\begin{equation}
    \label{block rate}
        BR=\frac{1}{N}\displaystyle\sum_{i=1}^{N}\mathbbm{1}\left [ \bigwedge_{(t,q)\in\mathcal{D}_{val}^{-}}(g(t,q_{i}) < 0.5)\right ],
\end{equation}

\noindent where $\mathcal{D}_{val}^{-}=\mathcal{D}_{val}-\left \{ (t_{i},q_{i})\right \}$. Intuitively, a higher value of $SR$ suggests that the scope detector is able to retrieve the desired edit statements for more atomic questions, while a higher value of $BR$ implies that fewer atomic questions are mistakenly categorized into the edit scope of the irrelevant edits. 
We use these two metrics to evaluate detectors $M_{\mathrm{pre}}$ and $M_{\mathrm{dis}}$, and return the optimal-performing detectors on validation set (i.e., having the highest sum of $SR$ and $BR$). We empirically find that these two metrics performs better serving as the indicator of early stopping \citep{earlystop}.

\subsection{Knowledge Prompt Generator}
\label{sub2}
To identify the leading subquestion during question decomposition, we propose knowledge prompt generator $M_{\mathrm{gen}}$, which aims to provide the additional valuable contextual information. Specifically, we employ ELQ~\citep{elq}, a fast end-to-end entity linking model. It recognizes the key entity, i.e., the named entity in the input question $Q$,  links the entity to Wikidata, 
and subsequently retrieves the related knowledge facts from Wikidata \citep{wikidata}. 

The retrieved knowledge facts from the Wikidata are the valuable contextual information for the question decomposition and the knowledge facts are stored as triplets $(s,r,o)$ in Wikidata. We adopt the following strategy to only preserve the commonsense facts from the vast knowledge base. For simplicity, we consider two \textit{basic membership properties} $\mathcal{R}=\left [ r_{1},r_{2}\right ]$ as our interested relations, where $r_{1}$=\textit{instance of}, $r_{2}$=\textit{subclass of}. Each entity in Wikidata possesses at least one of the relations. These two relations typically provide infallible commonsense facts related to the entity. Thus, for a key entity $s_{i}$, we randomly choose $(s_{i},r_{1},o_{1})$ or $(s_{i},r_{2},o_{2})$ as the retrieval fact. After retrieving, we use a manually-defined template to convert both key entity and retrieval fact into a knowledge prompt to augment the input question $Q$. For instance, as shown in Figure \ref{pokemqa_framework}, we recognize the key entity \textit{Messi} and retrieve the knowledge fact \textit{(Messi, instance of, human)}. After composing them together, we finally get the knowledge prompt \textit{Entity of Question: Messi, a human}.


\section{Experimental Setup}





We evaluate our approach on MQUAKE~\citep{mello}, which is a knowledge editing benchmark. It includes MQUAKE-CF-3K based on counterfactual edits, and MQUAKE-T with temporal knowledge updates. These datasets consist of a number of $k$-hop questions ($k \in \left \{ 2,3,4\right \}$), each of them is associated with one or more edits. More statistics can be found in Appendix \ref{statistics}.

\subsection{Evaluation Metrics}
\label{metric}

\noindent \textbf{Multi-hop accuracy}~\citep{mello}. It measures the accuracy of the (edited) language models in answering multi-hop questions.

\noindent \textbf{Hop-wise answering accuracy (Hop-Acc)}. In order to avoid the potential interference caused by unreliable reasoning, we propose the Hop-Acc to checks the correctness of intermediate reasoning path when evaluating MQA performance. Specifically, for a multi-hop question $Q$, since the question decomposition prompt is completely structured, language models are able to state the intermediate answer of subquestion in a concise, parseable way. Thus, the chain of intermediate answer $\left \langle s_{1}, o_{1},\dots ,o_{n}\right \rangle$ can be parsed from the inference content as the deduced path $\mathcal{P}$. We argue that a multi-hop question is fully solved by language models only if the deduced path $\mathcal{P}$ is exactly the same as the \textit{gold path} $\mathcal{P}^{*}$ (defined in Section \ref{hop-acc}), i.e., the novel metric measures the accuracy of reasoning path for multi-hop questions, which is only available for sequential question decomposition. 


\subsection{Baselines Methods \& Language Models}
We take four knowledge editing methods as baselines, including parameter updating methods, \textbf{FT} \citep{ft-zhu}, \textbf{ROME} \citep{rome}, \textbf{MEMIT} \citep{memit} and memory-based method \textbf{MeLLo} \citep{mello}. More implementation details are in Appendix \ref{baseline}. So far there is still no enough evidence to prove whether chain-of-thought (COT) prompting \citep{cot} or question decomposition (QD) prompting \citep{comgap} is more effective. Thus, to ensure fair and comprehensive comparisons, except for memory-based editors (MeLLo, PokeMQA) that relies on question decomposition, we report the performance of other parameter updating methods under both COT and QD prompting\footnote{These methods executes the entire process by itself, without the need for repeatedly prompting.}.

We conduct experiments on the following three base language models: 
\noindent\textbf{LLaMa-2-7B} \citep{llama2} is a powerful open-source pre-trained large language model, implemented by Huggingface Transformers library \citep{huggingface};
\noindent\textbf{Vicuna-7B} \citep{vicuna} is trained by fine-tuning LLaMA, implemented by Fastchat library \citep{fastchat};
\noindent\textbf{GPT-3.5-turbo-instruct} \citep{gpt-3.5} is a variant of the most capable GPT-3.5 series model, GPT-3.5-turbo (ChatGPT), which is used for legacy completion.

\begin{table*}[t]
\renewcommand\arraystretch{1.3}
\centering
\resizebox{0.98\linewidth}{!}{
    \footnotesize
\begin{tabular}{lcccccccccccc}
\toprule
\multicolumn{1}{c}{} &
   &
  \multicolumn{6}{c}{MQUAKE-CF-3K} &
   &
  \multicolumn{4}{c}{MQUAKE-T} \\ \cline{3-8} \cline{10-13} 
\multicolumn{1}{c}{} &
   &
  \multicolumn{2}{c|}{1 edited} &
  \multicolumn{2}{c|}{100 edited} &
  \multicolumn{2}{c}{All edited} &
   &
  \multicolumn{2}{c|}{1 edited} &
  \multicolumn{2}{c}{All edited} \\ \cline{3-8} \cline{10-13} 
\multicolumn{1}{c}{Method} &
   &
  Acc. &
  Hop-Acc &
  Acc. &
  Hop-Acc &
  Acc. &
  Hop-Acc &
   &
  Acc. &
  Hop-Acc &
  Acc. &
  Hop-Acc \\ \hline
\multicolumn{11}{c}{\textbf{LLaMa-2}} &
  \multicolumn{2}{r}{Size: 7B} \\ \hline
FT$_{\mathrm{COT}}$ &
   &
  22.3 &
  - &
  2.13 &
  - &
  OOM &
  - &
   &
  47.32 &
  - &
  3.75 &
  - \\
FT &
   &
  28.2 &
  7.3 &
  2.37 &
  0.03 &
  OOM &
  OOM &
   &
  56.48 &
  33.89 &
  1.02 &
  0.37 \\
ROME$_{\mathrm{COT}}$ &
   &
  11.17 &
  - &
  2.87 &
  - &
  2.77 &
  - &
   &
  28.96 &
  - &
  14.4 &
  - \\
ROME &
   &
  13.13 &
  5.37 &
  3.5 &
  0.03 &
  3.63 &
  0.1 &
   &
  24.89 &
  17.99 &
  1.71 &
  0.32 \\
MEMIT$_{\mathrm{COT}}$ &
   &
  11.83 &
  - &
  9.23 &
  - &
  5.57 &
  - &
   &
  36.88 &
  - &
  31.58 &
  - \\
MEMIT &
   &
  14.97 &
  6.43 &
  9.4 &
  2.47 &
  2.3 &
  0.37 &
   &
  30.89 &
  23.98 &
  25.21 &
  20.13 \\
MeLLo &
   &
  33.57 &
  9.9 &
  20.0 &
  10.07 &
  17.33 &
  9.9 &
   &
  \textbf{97.7} &
  0.21 &
  62.58 &
  3.96 \\ \hline
PokeMQA (Ours) &
   &
  \textbf{44.13} &
  \textbf{30.6} &
  \textbf{37.33} &
  \textbf{27.83} &
  \textbf{32.83} &
  \textbf{23.87} &
   &
  75.43 &
  \textbf{60.44} &
  \textbf{74.36} &
  \textbf{60.22} \\ \hline
\multicolumn{11}{c}{\textbf{Vicuna}} &
  \multicolumn{2}{r}{Size: 7B} \\ \hline
MeLLo &
   &
  22.7 &
  7.03 &
  12.83 &
  6.77 &
  10.9 &
  6.7 &
   &
  42.24 &
  1.12 &
  19.86 &
  1.28 \\
PokeMQA (Ours) &
   &
  \textbf{45.83} &
  \textbf{34.8} &
  \textbf{38.77} &
  \textbf{31.23} &
  \textbf{31.63} &
  \textbf{25.3} &
   &
  \textbf{74.57} &
  \textbf{55.19} &
  \textbf{73.07} &
  \textbf{55.09} \\ \hline
\multicolumn{11}{c}{\textbf{GPT-3.5-turbo-instruct}} &
  \multicolumn{2}{r}{Size: Undisclosed} \\ \hline
MeLLo &
   &
  57.43 &
  28.8 &
  40.87 &
  28.13 &
  35.27 &
  25.3 &
   &
  \textbf{88.12} &
  52.84 &
  74.57 &
  53.53 \\
PokeMQA (Ours) &
   &
  \textbf{67.27} &
  \textbf{56.37} &
  \textbf{56.0} &
  \textbf{49.63} &
  \textbf{45.87} &
  \textbf{39.77} &
   &
  76.98 &
  \textbf{68.09} &
  \textbf{78.16} &
  \textbf{67.88} \\ 
  \bottomrule
\end{tabular}
}

\caption{Evaluation results on MQUAKE-CF-3K and MQUAKE-T. The best result is indicated in \textbf{Bold}. The term `k edited' means the size of edit batch is k. `$\mathrm{COT}$' means that the current method uses chain-of-thought prompt, otherwise the question decomposition prompt; The metrics are multi-hop accuracy (Acc) and Hop-wise answering accuracy (Hop-Acc) presented in Section \ref{metric}. `-' means the current metric is not applicable to this setting.}

\label{main experiment}
\end{table*}

\subsection{Implementation Details}
We finetune the pre-detector $g_{\phi}$ and conflict disambiguator $g_{\psi}$ based on \textbf{DistilBERT} \citep{distilbert}. Note that the $\mathcal{D}_{\mathrm{train}}$ used for fine-tuning does not contain any edit statements $t$ that appear during testing. We provide detailed fine-tuning setting in Appendix \ref{hyperparameter}.

To evaluate performance under varying numbers of edits, we conduct stratified sampling \citep{stratifiedsampling} of the dataset according to hops of questions to construct edit batches of different sizes, which ensures the proportion of questions with different hops is relatively the same within each edit batch. We inject all the edits within a batch simultaneously\footnote{Since ROME is not able to perform batch edit, we sequentially inject edits within a batch.} \citep{easyedit}.

It should be noted that we conduct experiments related to parameter updating methods exclusively on the open-source LLM (\textbf{LLaMa-2-7B}), while the memory-based editing methods are comprehensively evaluated across all language models. (More details about experiments in Appendix \ref{detailexp}).

\section{Performance Analysis}
\subsection{Main Results}
\textbf{\underline{PokeMQA} is effective and reliable.    }We report our main results in Table \ref{main experiment}. The results demonstrate that PokeMQA outperforms all baselines by a large margin in almost all settings. Moreover, PokeMQA achieves the highest \textbf{Hop-Acc} across all settings, which strongly supports our view that the coupling of question decomposition and conflict detection places too much burden on LLMs, thus negatively impacting their inference abilities. PokeMQA significantly addresses the issue of unreliable reasoning, further improving MQA performance under knowledge editing. Meanwhile, achieving a high \textbf{Hop-Acc} indicates that PokeMQA's reasoning process is more rational and can serve as a more reliable explanation for model predictions, enhancing the interpretability of LLMs in MQA. Furthermore, PokeMQA can scale effectively with current mainstream LLMs, such as GPT-3.5-turbo-instruct, without the need for additional training.

\noindent \textbf{\underline{MeLLo} is a potential alternative.    }The related results about MeLLo suggest that it is undoubtedly a strong competitor. In the head-to-head comparisons on LLaMa-2-7B, MeLLo achieves (1/10) optimal result and (7/10) sub-optimal results. Surprisingly, MeLLo also achieves the best performance in two settings (On MQUAKE-T, when the size of edit batch is 1, 97.7 on LLaMa-2-7B and 88.12 using GPT-3.5-turbo-instruct). But through a detailed analysis of the reasoning processes, we discover that MeLLo resolves most multi-hop questions by exploiting \textit{shortcut reasoning patterns} (See an example in Appendix \ref{shortcutreasoningappen}), which can be considered as a form of underfitting to the prompt. A clear evidence is that the accuracy on LLaMa-2-7B with weaker inference ability is higher than on GPT-3.5-turbo. Meanwhile, it is undeniable that MeLLo's performance benefits significantly from the increased capabilities of LLMs, suggesting that on a stronger LLM in the future, MeLLo might further narrow the performance gap with PokeMQA.

\noindent \textbf{\underline{Parameter-updating} may not be the answer to knowledge editing.    }In line with previous researches (\citealp{mello}, \citealp{propaknowledge}), parameter updating methods fail catastrophically at answering multi-hop questions, indicating that the injected knowledge cannot be flexibly applied to inference by the edited model. FT achieves the best performance among these updating methods when the size of edit batch is 1, but the impact of a slightly larger edit batch on its performance can already be devastating. ROME performs worse than MEMIT in all settings, which is consistent with the fact that MEMIT is an improved version of ROME. MEMIT displays a certain level of robustness to the size of edit batch, but it still fails under thousands of edited facts. In summary, our results indicate that parameter updating methods can hardly meet the desiderata of knowledge editing applications.

\begin{figure*}[t]
    \centering
	\subfloat{
		\includegraphics[width=0.31\linewidth]{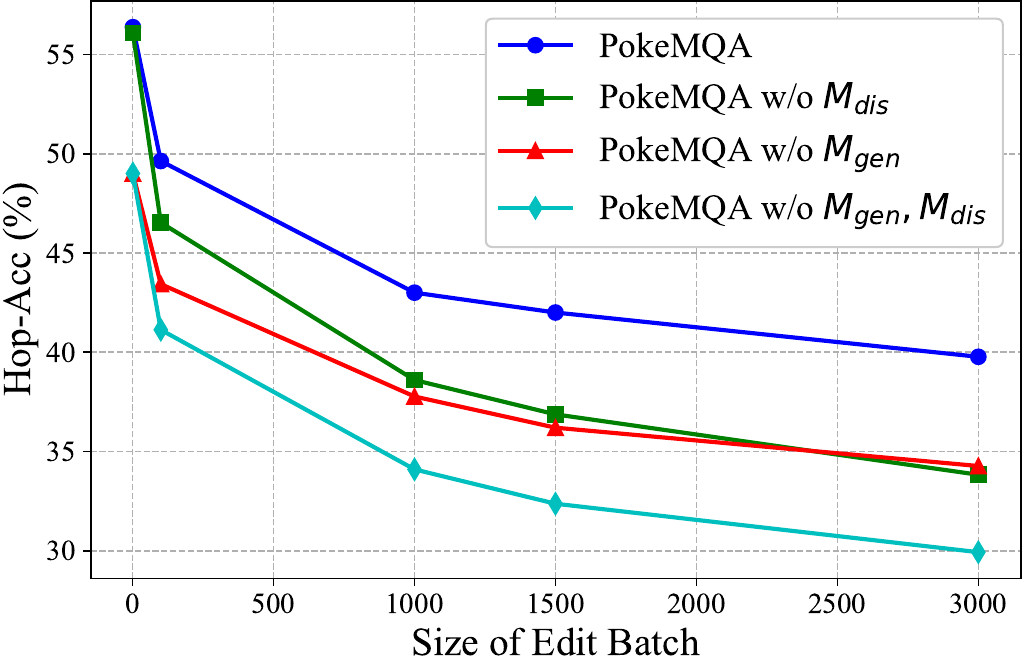}
	}
	\subfloat{
		\includegraphics[width=0.32\linewidth]{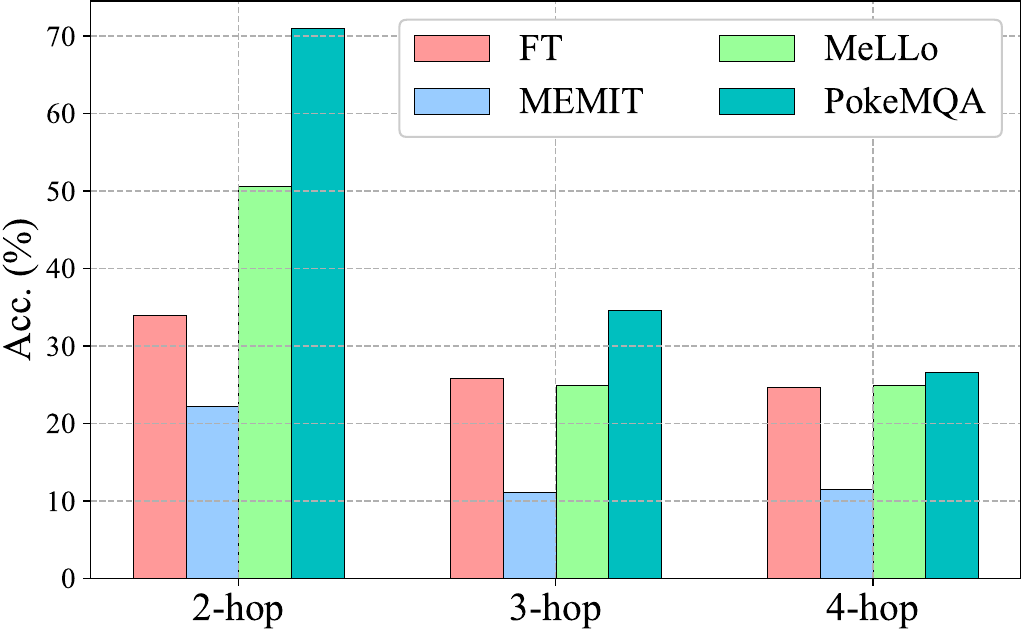}
	}
	\subfloat{
		\includegraphics[width=0.32\linewidth]{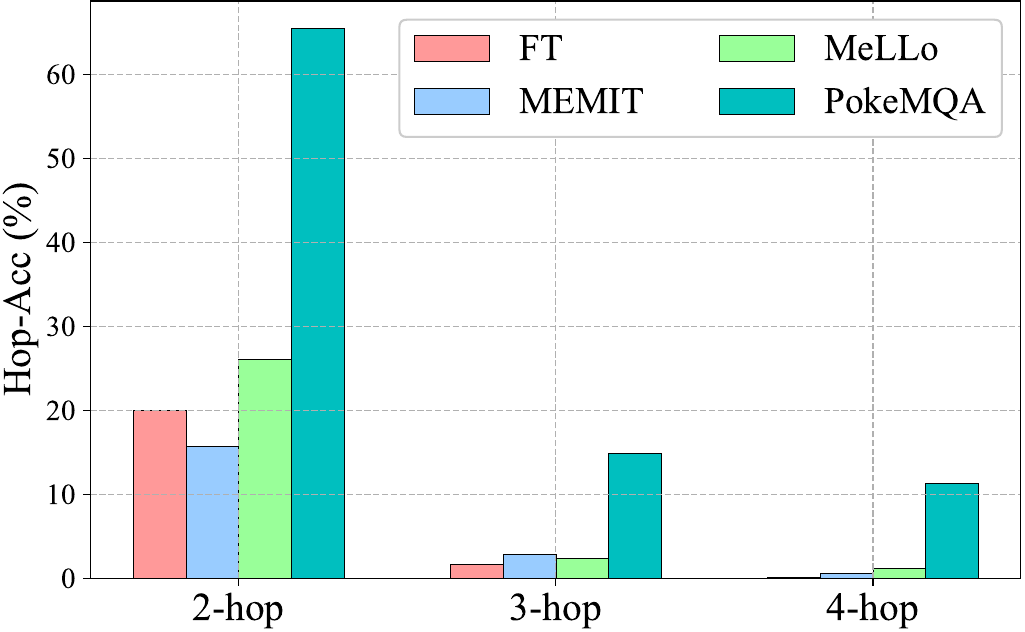}
	}
\caption{\textbf{Left}: Hop-Acc across multiple variants of PokeMQA with varying size of edit batch, utilizing GPT-3.5-turbo-instruct as the base language model on MQUAKE-CF-3K. \textbf{Middle, Right}: On MQUAKE-CF-3K, Acc. and Hop-Acc results for 2,3,4-hop questions, utilizing different knowledge editing methods. The experiments is conducted on LLaMa-2-7B with the size of edit batch is 1. Extra results is provided in Appendix \ref{promptinPMQA}.}
\label{mixfig}
\vspace{-2pt}
\end{figure*}

\noindent \textbf{\underline{MQA under knowledge editing} remains challenging.    } As shown in Figure \ref{mixfig} (Middle, Right), PokeMQA consistently maintains state-of-the-art performance across multi-hop questions of varying difficulty levels while significantly surpassing other competitors in producing reliable reasoning. But depressingly, the increasing difficulty of the questions also has a significant negative impact on PokeMQA's performance. Combining more facts to generate more complex reasoning processes poses a double challenge in terms of edited fact retrieval accuracy and language model reasoning capabilities. Currently, PokeMQA is not fully capable of addressing these challenges, indicating that this task remains challenging for future knowledge editing methods.


\begin{table*}[t]
\renewcommand\arraystretch{1.5}
\centering
\resizebox{0.96\linewidth}{!}{
    \footnotesize
\begin{tabular}{cc|cccc|cccc|cccc}
\toprule
\multicolumn{1}{l}{} &
  \multicolumn{1}{l|}{} &
  \multicolumn{4}{c|}{GPT-3.5-turbo-instruct} &
  \multicolumn{4}{c|}{LLaMa-2-7B} &
  \multicolumn{4}{c}{Vicuna-7B} \\ \cline{3-14} 
\multicolumn{2}{c|}{\textbf{$M_{\mathrm{dis}}$  $M_{\mathrm{gen}}$}} &
  \multicolumn{2}{c}{MQUAKE-CF-3K} &
  \multicolumn{2}{c|}{MQUAKE-T} &
  \multicolumn{2}{c}{MQUAKE-CF-3K} &
  \multicolumn{2}{c|}{MQUAKE-T} &
  \multicolumn{2}{c}{MQUAKE-CF-3K} &
  \multicolumn{2}{c}{MQUAKE-T} \\ \cline{3-14} 
 &
   &
  1 edited &
  All edited &
  1 edited &
  All edited &
  1 edited &
  All edited &
  1 edited &
  All edited &
  1 edited &
  All edited &
  1 edited &
  All edited \\ \hline
- &
  - &
  49.0 &
  29.93 &
  67.99 &
  55.67 &
  29.33 &
  19.47 &
  59.31 &
  52.19 &
  27.37 &
  16.43 &
  54.23 &
  48.39 \\
$\sqrt{}$ &
  - &
  49.0 &
  34.27 &
  \textbf{68.09} &
  67.77 &
  29.33 &
  22.87 &
  59.31 &
  59.1 &
  27.37 &
  19.37 &
  54.23 &
  54.12 \\
- &
  $\sqrt{}$ &
  56.07 &
  33.83 &
  68.04 &
  56.32 &
  \textbf{30.6} &
  20.3 &
  \textbf{60.44} &
  53.21 &
  \textbf{34.8} &
  22.23 &
  \textbf{55.19} &
  49.68 \\ \hline
$\sqrt{}$ &
  $\sqrt{}$ &
  \textbf{56.37} &
  \textbf{39.77} &
  \textbf{68.09} &
  \textbf{67.88} &
  \textbf{30.6} &
  \textbf{23.87} &
  \textbf{60.44} &
  \textbf{60.22} &
  \textbf{34.8} &
  \textbf{25.3} &
  \textbf{55.19} &
  \textbf{55.09} \\
\bottomrule
\end{tabular}
}
\caption{Ablation results of PokeMQA and its variants in terms of Hop-Acc. We also provide the results in terms of Acc. in Appendix \ref{promptinPMQA}.}
\label{ablation experiment}
\vspace{-10pt}
\end{table*}

\subsection{Ablation Study}
We conduct ablation experiments to investigate how the two detachable components $M_{\mathrm{dis}}$ and $M_{\mathrm{gen}}$ improves PokeMQA and analyze their necessity. The results are shown in Table \ref{ablation experiment} and Figure \ref{mixfig} (Left). Based on the experimental results, we find that the two components indeed enhance PokeMQA and summarize two conclusive findings as follows:


\noindent\textbf{Use $M_{\mathrm{gen}}$ selectively. } As shown in Table \ref{ablation experiment}, Figure \ref{mixfig} (Left), the knowledge prompt generator $M_{\mathrm{gen}}$ improves PokeMQA performance in almost all settings. Although the above results verify its effectiveness, we have to point out that the performance gain is much more significant in MQUAKE-CF-3K compared to MQUAKE-T. Our view is that this is because MQUAKE-T is constructed based on real fact updates in recent years, so the key entity in the input question may be familiar to the latest pre-trained LLMs. Consequently, they can recognize the key entity and access entity-related knowledge relatively easily, even without additional contextual information. Due to the extra computation cost of $M_{\mathrm{gen}}$, we recommend using $M_{\mathrm{gen}}$ selectively depending on the specific application.

\noindent\textbf{$M_{\mathrm{dis}}$ is indispensable for large-scale editing.} As shown in Table \ref{ablation experiment}, Figure \ref{mixfig} (Left), $M_\mathrm{dis}$ can bring more performance gains when edit batch size is larger. Meanwhile, we calculate the average number of predictions of $M_{\mathrm{dis}}$ (2.746 in MQUAKE-CF-3K and 4.565 in MQUAKE-T both on LLaMa-2-7B when the size of edit batch is \textbf{all}). The results indicate that by incorporating the $M_{\mathrm{dis}}$, a tiny additional number of predictions greatly boosts MQA performance under large edit batch. We conclude that $M_{\mathrm{dis}}$ can greatly enhance the robustness of PokeMQA to large-scale editing with almost no additional computational cost, which is indispensable to maintain the applicability of PokeMQA in real scenarios.

\section{Related Work}
\noindent\textbf{Knowledge editing methods.}  Knowledge editing focuses on updating factual knowledge to language models and a lot of related research has been carried out. Most of these methods predict updates to the weights of the base model by knowledge locating or meta-learning and then locally modify parameters (\citealp{mend}, \citealp{memit}). Another part preserves parameters and explicitly stores edit instances (\citealp{serac}, \citealp{mello}). 
Recent work has identified the limitations of existing editing methods through theoretical analysis \citep{representationdenoising} and performance evaluation \citep{propaknowledge}. Our work focuses on addressing one of these challenging tasks: MQA under knowledge editing scenarios.

\section{Conclusion}
In this work, we propose a novel programmable knowledge editing method (PokeMQA) to improve MQA performance and address unreliable reasoning. PokeMQA leverages a scope detector to align LLMs' behavior with edited facts and incorporates auxiliary knowledge prompt to enrich contextual information. Extensive experiments across three LLMs show that PokeMQA helps LLMs answer multi-hop questions in a precise and reliable way.
\section*{Limitations}
In this work, we did not design a task-specific architecture for the scope detector to achieve higher fact retrieval accuracy and mitigate the pressure that the context length imposes on the LLMs reasoning capabilities when handling complex multi-hop questions.

Besides, although memory-based editing shows great potential for controlled editing and large-scale editing, the way it stores edit instances makes it extremely vulnerable to attacks such as memory injection. Therefore, memory-based editing needs to be supported by reliable security technology to reduce its risk in real scenarios.

\bibliography{custom}

\appendix
\section{Prompt \& Supplement Results for PokeMQA}
\label{promptinPMQA}
\begin{figure*}[ht]
    \centering
    \subfloat[GPT-3.5-turbo-instruct (Acc.)]{
		\includegraphics[width=0.33\linewidth]{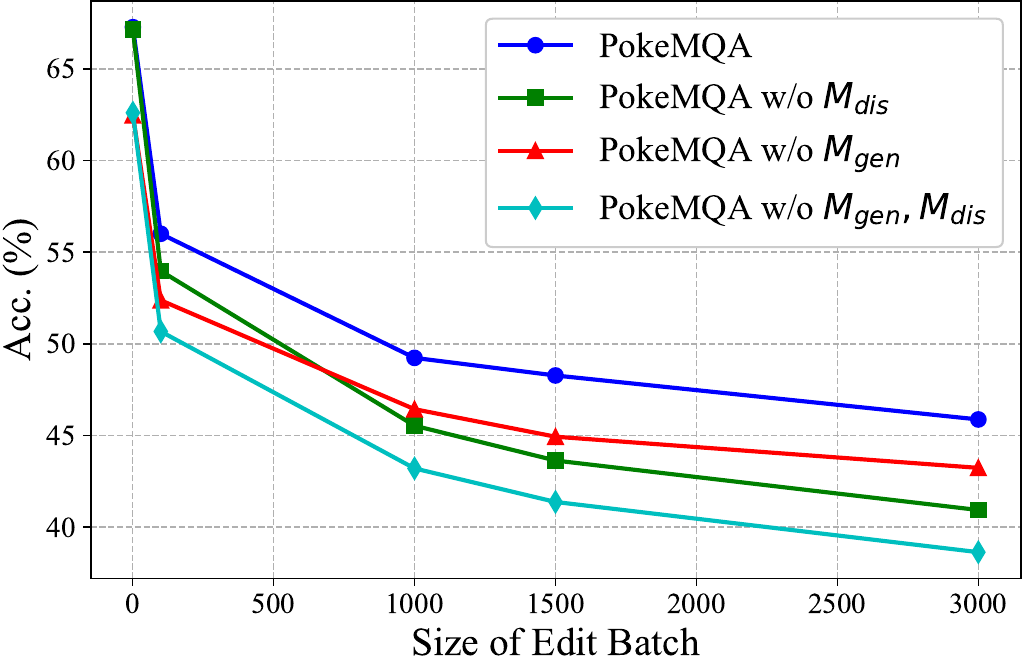}
	}
	\subfloat[LLaMa-2-7B (Hop-Acc)]{
		\includegraphics[width=0.33\linewidth]{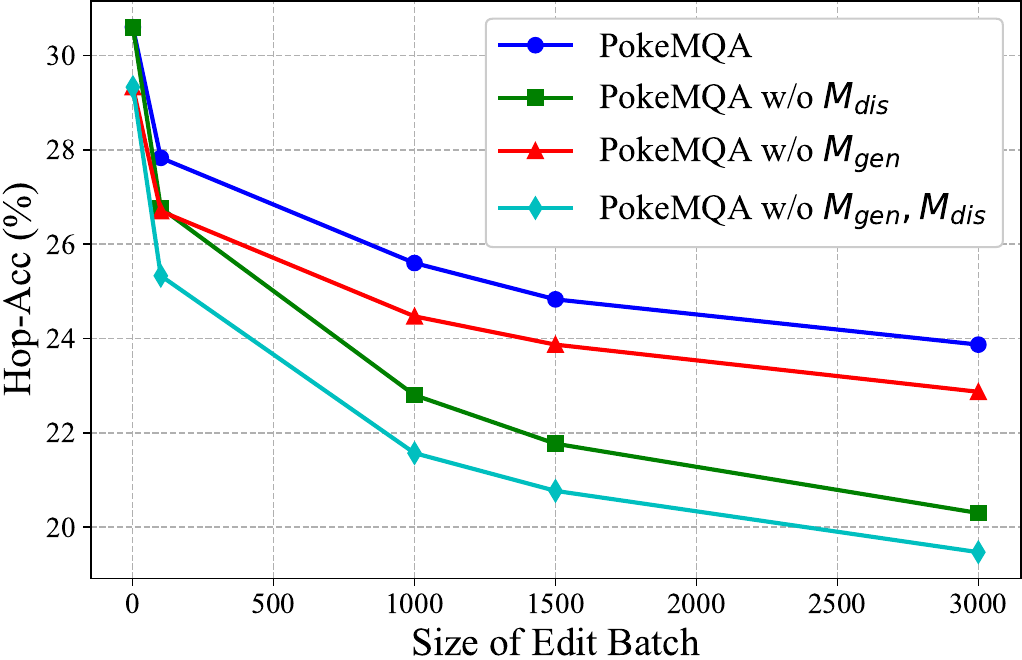}
	}
    \subfloat[LLaMa-2-7B (Acc.)]{
		\includegraphics[width=0.33\linewidth]{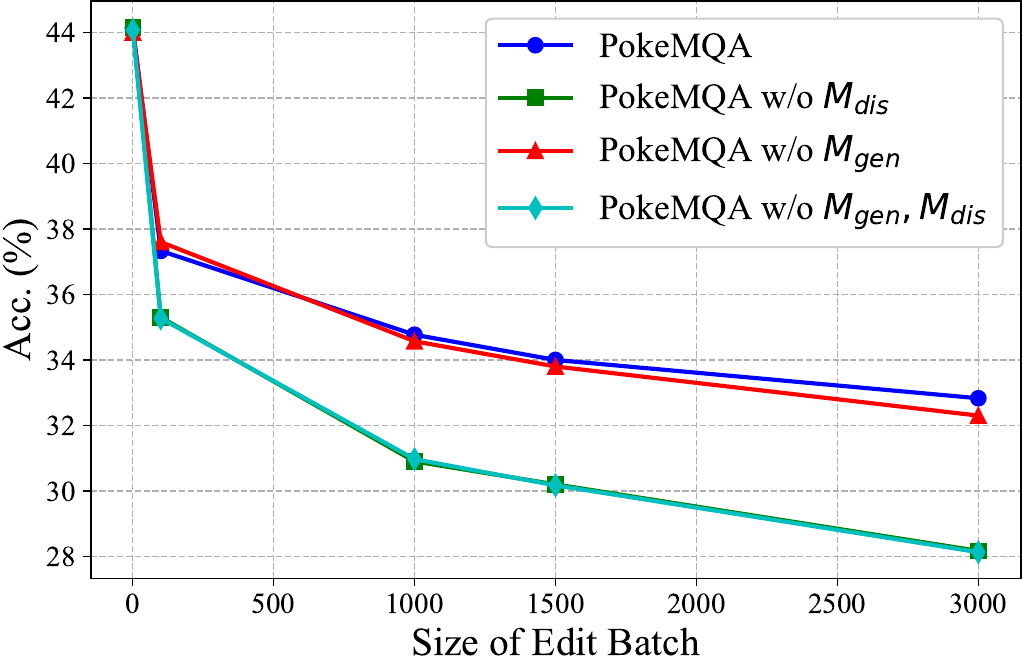}
	}
\caption{Hop-Acc and Acc. across multiple variants of PokeMQA in MQUAKE-CF-3K on GPT-3.5-turbo-instruct and LLaMa-2-7B with edit batches of different sizes.}
\label{curve-remaining}
\end{figure*}

A demonstration example used in prompt is shown in Table \ref{PokeMQA-prompt}. The supplement ablation results are shown in Table \ref{abl-acc} and Figure \ref{curve-remaining}.
\section{Details of Training Dataset Construction}
\label{datasetconstruction}

To train our scope detector, including pre-detector $g_{\phi}$ and conflict disambiguator $g_{\psi}$, we construct a training dataset $\mathcal{D}$. Specifically, we first extract edit triples from MQUAKE-CF and filter out the part sharing the same $(s,r)$ with fact triples appeared in MQUAKE-CF-3K and MQUAKE-T, constructing a edit dataset $\mathcal{D}_{e}=\left \{ e_{1},\dots ,e_{n}\right \}$. Then we use manually-defined template to convert each edit triple $e$ into a natural language statement $s$ and get a edit statement dataset $\mathcal{D}_{state}=\left \{ t_{1},\dots ,t_{n}\right \}$. Finally we design a prompt consisting of instruction and demonstrations (shown in Table \ref{dataconstructprompt}) and prompt Vicuna-13B \cite{vicuna} to generate three diversely phrased atomic questions for each $s \in \mathcal{D}_{state}$, building a training dataset $\mathcal{D}=\left \{ \left ( t_{1},q_{1}\right ),\dots ,\left ( t_{n},q_{n}\right )\right \}$. It should be noted that when computing $SR$ and $BR$, or sampling negative instances, instances with the same statement $t$ should not be considered.

\section{Multi-hop Question Answering Dataset Statistics}
\label{statistics}
Table \ref{datasetstatistic} contains the statistics for the two benchmark datasets used in our experiments.

\begin{table}[h]
\resizebox{0.98\linewidth}{!}{
    \footnotesize
\begin{tabular}{llcccc}
\toprule
 & \textbf{\#Edits} & \multicolumn{1}{l}{\textbf{2-hop}} & \multicolumn{1}{l}{\textbf{3-hop}} & \multicolumn{1}{l}{\textbf{4-hop}} & \multicolumn{1}{l}{\textbf{Total}} \\ \hline
                                 & 1       & 513  & 356  & 224  & 1093 \\
                                 & 2       & 487  & 334  & 246  & 1067 \\
\multicolumn{1}{c}{MQUAKE-CF-3K} & 3       & -    & 310  & 262  & 572  \\
                                 & 4       & -    & -    & 268  & 268  \\
                                 & All     & 1000 & 1000 & 1000 & 3000 \\ \hline
\multicolumn{1}{c}{MQUAKE-T}     & 1 (All) & 1421 & 445  & 2    & 1868 \\
\bottomrule
\end{tabular}
}
\caption{Statistics of datasets used in experiments}
\label{datasetstatistic}
\end{table}

Besides, there are 2786 different edits for MQUAKE-CF-3k and 96 different edits for MQUAKE-T when the size of edit batch is \textbf{all}.

\section{Details about Scope Detector finetuning}
\label{hyperparameter}
We finetune the pre-detector $g_{\phi}$ and conflict disambiguator $g_{\psi}$ based on \textbf{DistilBERT} \citep{distilbert} and the checkpoint is \textit{distilbert-base-cased} from Huggingface Transformers library \citep{huggingface}. We take $SR+BR-1$ as the indicator of early stopping.

To fine-tune pre-detector $g_{\phi}$, the
learning rate is set as $1e^{-5}$ with Adam optimizer \cite{adam}, the batch size is set as 1024, and the number of negative samples is 20; To fine-tune conflict disambiguator $g_{\psi}$, the
learning rate is set as $1e^{-5}$ with Adam optimizer, the batch size is set as 256 and the number of negative samples is 1. And the dataset split is 80\%/20\% for training and validation, without the need of testing.

\section{Implementation Details of Baselines}
\label{baseline}
In our experiments, the parameter updating knowledge editing methods, including \textbf{FT}, \textbf{ROME} and \textbf{MEMIT} is implemented by EasyEdit library \cite{easyedit}. We basically follow the default hyperparameter settings on \textbf{LLaMa-2-7B} in library, and make a slight adjustment to ensure the effectiveness of these methods in different experimental settings. We modify the learning rate for ROME and target editing layer for MEMIT, the detailed modifications is shown in Table \ref{baselinehyperpara}.

\begin{table}[h]
    \centering
\resizebox{0.98\linewidth}{!}{
    \footnotesize
    \begin{tabular}{clccccc}
    \toprule
\textbf{} &  & \multicolumn{3}{c}{MQUAKE-CF-3K}                        & \multicolumn{2}{c}{MQUAKE-T} \\ \cline{3-7} 
Method    &  & 1 edited  & \multicolumn{1}{l}{100 edited} & All edited & 1 edited     & All edited    \\ \hline
ROME      &  & $5e^{-1}$ & $5e^{-5}$                      & $5e^{-6}$  & $5e^{-1}$    & $1.5e^{-1}$   \\
MEMIT     &  & 4,5,6,7,8 & 5,6,7                          & 7          & 4,5,6,7,8    & 4,5,6,7,8    
\\
\bottomrule
\end{tabular}
}
    \caption{Detailed hyperparameter modification for ROME and MEMIT.}
    \label{baselinehyperpara}
\end{table}

\section{Two-stage edited fact Retrieval}
\label{twostage}
\begin{algorithm}[h]
	\renewcommand{\algorithmicrequire}{\textbf{Input:}}
	\renewcommand{\algorithmicensure}{\textbf{Output:}}
	\caption{Two-stage Edited Fact Retrieval.}
	\label{alg:1}
	\begin{algorithmic}[1]
		\REQUIRE edited memory $\mathcal{M}$, pre-detector $g_{\phi}$, conflict disambiguator $g_{\psi}$, atomic question $q$

        \STATE Initialize candidate set $\mathcal{Z}=\varnothing$
        \STATE Initialize final set $\mathcal{F}=\varnothing$
        \STATE {\color{blue}/* Pre-detection stage */}
		\FORALL{$ t_{i} \in \mathcal{M}$}
		\STATE \textbf{if} $g_{\phi}(t_{i},q) \ge 0.5$ \textbf{then} $\mathcal{Z}=\mathcal{Z}\bigcup \left \{t_{i} \right \}$
		\ENDFOR
        \IF{$\left | \mathcal{C}\right |= 1$}
        \STATE \textbf{return} $t_{i^{*}}$, where $t_{i^{*}} \in \mathcal{Z}$
        \ENDIF
        \STATE {\color{blue}/* Conflict-disambiguation stage */}
        \FORALL{$ t_{i} \in \mathcal{C}$}
		\STATE \textbf{if} $g_{\psi}(t_{i},q) \ge 0.5$ \textbf{then} $\mathcal{F}=\mathcal{F}\bigcup \left \{t_{i} \right \}$
		\ENDFOR
        \IF{$\mathcal{F} \neq \varnothing $}
        \STATE \textbf{return} $t_{i^{*}}$, where $i^{*}=\arg \max_{i}g_{\psi}(t_{i},q)$, $t_{i} \in \mathcal{F}$
        \ENDIF
	\end{algorithmic} 
\end{algorithm}

\section{Licensing}
Vicuna-7B (v1.1) and distilbert-base-cased are released under the Apache License 2.0. LLaMa-2-7B is licensed under the LLAMA 2 Community License. ELQ, ROME, MEMIT, FT are released under the MIT license.

\section{Details about Experiments}
\label{detailexp}
Because MQUAKE regard a \textit{chain of facts} as an instance and there are three diversely phrased multi-hop questions $Q$ for each instance, we follow \citep{mello}, if any of the three questions is considered solved in terms of the specific metric, the instance is considered correct.

The experiments, data, language models in the paper are all in English. We run all experiments on a machine with four
NVIDIA A40 GPU. One run of our experiments takes about 15 GPU hours. For all experiments, we use greedy decoding strategy to get the output in text space of language models for reproducibility and report a single run result due to the
limited computational resources.

\section{Shortcut Reasoning}
\label{shortcutreasoningappen}
Given an example from Table \ref{shortcut-reasoning}, although MeLLo appears to successfully combine two facts to arrive at the final answer, its reasoning process does not adhere to the task logic demonstrated in the prompt, which can be regarded as an underfitting to question decomposition.

\begin{table*}[t]
\centering
\resizebox{0.98\linewidth}{!}{
    \footnotesize
\begin{tabular}{|lll|}
\hline
\multicolumn{3}{|l|}{\begin{tabular}[c]{@{}l@{}}Question: What is the capital city of the country of citizenship of Ivanka Trump's spouse?\\ Entity of Question: \sethlcolor{c1}\hl{Ivanka Trump, a human.}\\ Subquestion: Who is Ivanka Trump's spouse?\\ Generated answer: Ivanka Trump's spouse is Jared Kushner.\\ According to Generated answer, the entity of Subquestion is: Jared Kushner\\ Subquestion: What is the country of citizenship of Jared Kushner?\\ Generated answer: \sethlcolor{c2}\hl{Jared Kushner is a citizen of Canada.}\\ According to Generated answer, the entity of Subquestion is: Canada\\ Subquestion: What is the capital city of Canada?\\ Generated answer: The capital city of Canada is Ottawa.\\ According to Generated answer, the entity of Subquestion is: Ottawa\\ Final answer: Ottawa\end{tabular}} \\ \hline
\end{tabular}
}
\caption{A in-context demonstration example used in our PokeMQA prompt, here we omit the remaining three demonstrations. \sethlcolor{c1}\hl{This color} indicate that this part is constructed after being retrieved by knowledge prompt generator from external knowledge base. \sethlcolor{c2}\hl{This color} indicate that this part is retrieved by scope detector from external memory.}
\label{PokeMQA-prompt}
\end{table*}

\begin{table*}[t]
\renewcommand\arraystretch{1.5}
\centering
\resizebox{0.98\linewidth}{!}{
    \footnotesize
\begin{tabular}{cc|cccc|cccc|cccc}
\toprule
\multicolumn{1}{l}{} &
  \multicolumn{1}{l|}{} &
  \multicolumn{4}{c|}{GPT-3.5-turbo-instruct} &
  \multicolumn{4}{c|}{LLaMa-2-7B} &
  \multicolumn{4}{c}{Vicuna-7B} \\ \cline{3-14} 
\multicolumn{2}{c|}{\textbf{$M_{\mathrm{dis}}$  $M_{\mathrm{gen}}$}} &
  \multicolumn{2}{c}{MQUAKE-CF-3K} &
  \multicolumn{2}{c|}{MQUAKE-T} &
  \multicolumn{2}{c}{MQUAKE-CF-3K} &
  \multicolumn{2}{c|}{MQUAKE-T} &
  \multicolumn{2}{c}{MQUAKE-CF-3K} &
  \multicolumn{2}{c}{MQUAKE-T} \\ \cline{3-14} 
 &
   &
  1 edited &
  All edited &
  1 edited &
  All edited &
  1 edited &
  All edited &
  1 edited &
  All edited &
  1 edited &
  All edited &
  1 edited &
  All edited \\ \hline
- &
  - &
  62.6 &
  38.63 &
  76.87 &
  65.63 &
  44.07 &
  28.13 &
  74.68 &
  65.90 &
  44.03 &
  26.27 &
  73.72 &
  66.27 \\
$\sqrt{}$ &
  - &
  62.47 &
  43.23 &
  \textbf{77.09} &
  \textbf{78.53} &
  44.0 &
  32.3 &
  74.68 &
  73.82 &
  44.03 &
  29.87 &
  73.72 &
  72.38 \\
- &
  $\sqrt{}$ &
  67.13 &
  40.93 &
  76.93 &
  66.06 &
  \textbf{44.17} &
  28.17 &
  \textbf{75.43} &
  66.6 &
  \textbf{45.9} &
  28.2 &
  \textbf{74.57} &
  67.29 \\ \hline
$\sqrt{}$ &
  $\sqrt{}$ &
  \textbf{67.27} &
  \textbf{45.87} &
  76.98 &
  78.16 &
  44.13 &
  \textbf{32.83} &
  \textbf{75.43} &
  \textbf{74.36} &
  45.83 &
  \textbf{31.63} &
  \textbf{74.57} &
  \textbf{73.07}
  \\
  \bottomrule
\end{tabular}
}
\caption{Ablation study results of PokeMQA and its variants in terms of Acc.}
\label{abl-acc}
\end{table*}

\begin{table*}[t]
\centering
\begin{tabular}{|lll|}
\hline
\multicolumn{3}{|l|}{Please generate three different phrased Questions for each fact} \\
 &
   &
   \\
\multicolumn{3}{|l|}{\begin{tabular}[c]{@{}l@{}}Fact: The univeristy where Bob Dylan was educated is University of Minnesota.\\ Question 1: What is the name of the educational institution where Bob Dylan studied?\\ Question 2: What is the name of the university where Bob Dylan was educated?\\ Question 3: At which university did Bob Dylan receive his education?\end{tabular}} \\
 &
   &
   \\
\multicolumn{3}{|l|}{\begin{tabular}[c]{@{}l@{}}Fact: The capital of United Kingdom is Angri.\\ Question 1: What is the name of the capital city of United Kingdom?\\ Question 2: Which city serves as the capital of United Kingdom?\\ Question 3: In which city is the capital of United Kingdom located?\end{tabular}} \\
 &
   &
   \\
\multicolumn{3}{|l|}{\begin{tabular}[c]{@{}l@{}}Fact: basketball was created in the country of Spain.\\ Question 1: Where is basketball originated from?\\ Question 2: What is the name of the country of origin of basketball?\\ Question 3: Where did the sport of basketball originate?\end{tabular}} \\
 &
   &
   \\
\multicolumn{3}{|l|}{\begin{tabular}[c]{@{}l@{}}Fact: John Coltrane is married to Anne Hathaway.\\ Question 1: Who is John Coltrane's spouse?\\ Question 2: To whom is John Coltrane married?\\ Question 3: Which person is married to John Coltrane?\end{tabular}} \\
 &
   &
   \\
\multicolumn{3}{|l|}{\begin{tabular}[c]{@{}l@{}}Fact: Miranda Kerr is a citizen of Australia.\\ Question 1: Which country is Miranda Kerr a citizen of?\\ Question 2: What is the name of the country that Miranda Kerr belongs to?\\ Question 3: What is Miranda Kerr's nationality?\end{tabular}} \\ \hline
\end{tabular}
\caption{The rephrased question generation prompt we used for dataset construction. The prompt contains one instruction and some in-context demonstrations.}
\label{dataconstructprompt}
\end{table*}

\begin{table*}[t]
\centering
\resizebox{0.98\linewidth}{!}{
    \footnotesize
\begin{tabular}{|l|}
\hline
Question: What is the name of the current head of state of Buckcherry's country of origin?       \\
  \\
Subquestion: What is the country of origin of the band Buckcherry?                               \\
                        \\
Generated answer: Buckcherry's country of origin is United States of America.                    \\
                        \\
Retrieved fact: The name of the current head of state in United States of America is Joe Biden.  \\
                        \\
Retrieved fact does not contradict to generated answer, so the intermediate answer is: Joe Biden \\
                        \\
Final answer: Joe Biden \\ \hline
\end{tabular}
}
\caption{An example of shortcut reasoning that occurs in the inference process of GPT-3.5-turbo-instruct. }
\label{shortcut-reasoning}
\end{table*}

\end{document}